**Informative Title**:

A Deep Dive into Understanding Tumor Foci Classification using Multiparametric MRI Based on Convolutional Neural Network

**Running Title**:

PCa MRI Interpretation Via Deep Learning

**Author list**:

Weiwei Zong, Ph.D.[1], Joon K. Lee, M.D.[1], Chang Liu, Ph.D.[1], Eric N. Carver [1,2], Aharon M. Feldman, M.D.[1], Branislava Janic, Ph.D.[1], Mohamed A. Elshaikh, M.D.[1], Milan V. Pantelic, M.D.[3], David Hearshen, M.D.[3], Indrin J. Chetty, Ph.D.[1], Benjamin Movsas, M.D.[1], Ning Wen, Ph.D.[1].

[1]Department of Radiation Oncology, Henry Ford Health System, Detroit, MI, USA

[2]School of Medicine Medical Physics, Wayne State University, Detroit, MI, USA

[3]Department of Radiology, Henry Ford Health System, Detroit, MI, USA

**Corresponding author**:

Corresponding Author: Ning Wen

Address: Department of Radiation Oncology, Henry Ford Hospital, 2799 W. Grand Blvd, Detroit, MI 48202

Phone: (313)916-3061

Fax: (248)661-7164

Email: nwen1@hfhs.org

# Abstract

**Purpose:** Deep learning models have had a great success in disease classifications using large data pools of skin cancer images or lung X-rays. However, data scarcity has been the roadblock of applying deep learning models directly on prostate multiparametric MRI (mpMRI). Although model interpretation has been heavily studied for natural images for the past few years, there has been a lack of interpretation of deep learning models trained on medical images.

In this paper, an efficient convolutional neural network (CNN) was developed and the model interpretation at various convolutional layers was systematically analyzed to improve the understanding of how CNN interprets multimodality medical images and the predictive powers of features at each layer. The problem of small sample size was addressed by feeding the intermediate features into a traditional classification algorithm known as weighted extreme learning machine (wELM), with imbalanced distribution among output categories taken into consideration.

**Methods:** The training data collection used a retrospective set of prostate MR studies, from SPIE-AAPM-NCI PROSTATEx Challenges held in 2017. Three hundred twenty biopsy samples of lesions from 201 prostate cancer patients were diagnosed and identified as clinically significant (malignant) or not significant (benign). All studies included T2-weighted (T2W), proton density-weighted (PD-W), dynamic contrast enhanced (DCE) and diffusion-weighted (DW) imaging. After registration and lesion-based normalization, a CNN with four convolutional layers were developed and trained on 10-fold cross validation. The features from intermediate layers were then extracted as input to wELM to test the discriminative power of each individual layer. The best performing model from the 10 folds was chosen to be tested on the holdout cohort from two sources. Feature maps after each convolutional layer were then visualized to monitor the trend, as the layer propagated. Scatter plotting was used to visualize the transformation of data distribution. Finally, a class activation map was generated to highlight the region of interest based on the model perspective.

**Results:** Experimental trials indicated that the best input for CNN was a modality combination of T2W, apparent diffusion coefficient (ADC) and $DWI_{b50}$. The convolutional features from CNN paired with a weighted extreme learning classifier showed substantial performance compared to a CNN end-to-end training model. The feature map visualization reveals similar findings on natural images where lower layers tend to learn lower level features such as edges, intensity changes, etc, while higher layers learn more abstract and task-related concept such as the lesion ~~parts~~ region. The generated saliency map revealed that the model was able to focus on the region of interest where the lesion resided and filter out background information, including prostate boundary, rectum, etc.

**Conclusions:** This work designs a customized workflow for the small and imbalanced data set of prostate mpMRI where features were extracted from a deep learning model and then analyzed by a traditional machine learning classifier. In addition, this work contributes to revealing how deep learning models interpret mpMRI for prostate cancer patients stratification.

**Keywords**: prostate cancer mpMRI lesion classification, small sample size, convolutional neural network, model interpretation, saliency map.

# 1. Introduction

Prostate cancer (PCa) is the most common cancer in men in the United States, and it is the second leading cause of cancer death in these patients[1]. Multiparametric magnetic resonance imaging (mpMRI) for PCa diagnosis has been increased significantly over the past decade. mpMRI sequences have shown promise for the detection and localization of PCa including T2-weighted (T2W), diffusion-weighted imaging (DWI), dynamic contrast-enhanced imaging (DCE) and MR spectroscopy. Combining these MR sequences into a multiparametric format has improved the performance characteristics of PCa detection and localization by evaluating area under curve (AUC) values, sensitivities, specificities, and positive predictive values[2,3]. Prostate biopsy is still considered the golden standard to determine if a suspicious lesion is benign or malignant. However, biopsy is an invasive procedure prone to complications such as hemorrhage, dysuria, and infection. Furthermore, in a small number of cases prostate biopsies can fail to establish the diagnosis despite magnetic resonance- and transrectal ultrasound-guided approaches[4].

Clinically significant cancer is defined on histopathology as Gleason score $\geq 7$ (including 3+4) according to PI-RADS™ v2 in order to standardize reporting of mpMRI and correlate imaging findings with pathology results[5]. Also, PCa is a multifocal disease in up to 87% of cases; therefore, the ability to distinguish malignant from benign foci within the prostate is crucial for optimal diagnosis and treatment. This has led to an interest in machine learning and computer vision utilizing mpMRI to non-invasively obtain accurate radiologic diagnoses that correlate with their histopathologic variants[6]. Unfortunately, data scarcity is one of the major challenges in applying deep machine learning algorithms in interpreting mpMRI images due to tightly regulated information acquisition as well as the high cost of MRI acquisition and data labelling from medical experts.  In addition, health records cannot be shared without consent due to privacy laws and related healthcare policies and regulations. Recently, efforts have been made[7,8] to transfer knowledge from publicly available large-scale data consisting of millions or more natural objects or scene images[9] to medical applications with a small scale of data. Unfortunately, limited improvements have been achieved yet mainly due to the remarkable differences between natural and medical images.

Among works studying deep learning models on prostate MRI, the evaluation of cross-modality interaction has rarely been touched. Compared to the recent progress in the application of deep neural networks on 2D images such as chest x-ray[10], mammogram[11], the challenges on the classification using prostate MRI[12,13] include relative small patient sample sizes due to high screening cost and much more complex information such as morphological, diffusion or perfusion imaging characteristics.

In this paper we report a refined convolutional neural network (CNN) with an appropriate depth coupled with a weighted extreme learning machine (wELM) classification method that has been carefully structured to adapt to a small data of 201 patients and 320 lesions. The purposes of this study are: 1) to test how a traditional shallow classifier can help improve the accuracy of the deep learning model given a small sample size; 2) to provide insight how the cross-modality information from mpMRI is used to characterize intraprostatic lesions in the deep learning model; 3) to understand how the CNN interprets prostate mpMRI sequences at each convolutional layer by visualizing feature maps and attention maps in the process of lesion classification.

# 2. Materials and Methods

## 2.1 Data

Data used in the study originated from the SPIE-AAPM-NCI Prostate MR Gleason Grade Group Challenge[14], which aimed to develop quantitative mpMRI biomarkers for determination of malignant lesions in PCa patients. PCa patients were previously de-identified by SPIE-AAPM-NCI and The Cancer Imaging Archive (TCIA). The images were acquired on two different types of Siemens 3T MR scanners, the MAGNETOM Trio and Skyra. T2W images were acquired using a turbo spin echo sequence and had a resolution of around 0.5 mm in plane and a slice thickness of 3.6 mm. The DWI series were acquired with a single-shot echo planar imaging sequence with a resolution of 2 mm in-plane and 3.6 mm slice thickness and with diffusion-encoding gradients in three directions. Three b-values were acquired (50, 400, and 800 s/mm$^2$), and subsequently, the ADC map was calculated by the scanner software[14].

The training set consisted of 320 lesion findings from 201 patients There was an independent testing cohort (test cohort 1) of 141 patients with 208 lesions used for a blind test of the model performance.

In the four examples shown in Figure 1, two were malignant lesions while the other two were benign. In the second case (row 2), there was a suspicious lesion in the peripheral zone (PZ) which showed hypointense signal in both T2W and ADC. Although the $k^{trans}$ was elevated in the prostate, the hyperintense signal was shown in the central gland (CG) instead. Similarly, the suspicious lesion of the fourth case (row 4) was noticeable in the CG for in four imaging modalities, however, it was classified as the benign lesion based on biopsy results while the third case was considered as malignant.

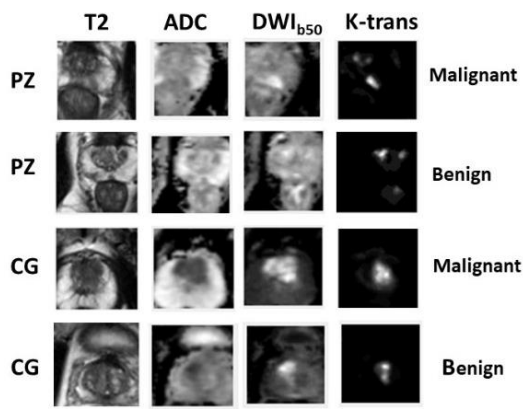

Figure 1. Display of T2, ADC, $DWI_{b50}$ and $k^{trans}$ of 2 malignant and 2 benign 64 x 64 patches including lesions residing in the peripheral zone (PZ) and central gland (CG), respectively. Pitfalls of each sequence can be observed as further illustrated in the text. Abbreviations: PZ – peripheral zone; CG – central gland.

To test cross-institutional generalization capability of our model, an independent cohort (test cohort 2) consisting of 25 patients with 33 lesions was collected for testing from our own institution with the IRB-approval. An ultrasound-guided needle biopsy was performed to confirm the diagnosis. Among the 33 lesions, 15 were diagnosed malignant and 18 benign. Two image modalities were acquired for each patient using a 3.0 T MR scanner (Ingenia, Philips Medical System, Best, the Netherlands) using small field of view as follows: T2W acquired with Fast-Spin-Echo (TE/TR: 4389/110ms, Flip Angle: 90º with image resolution of 0.42×0.42×2.4mm$^3$) and DWI with two b values (0 and 1,000 s/mm$^2$). The voxel-wise ADC map was constructed using these two DWIs.

All images were registered to T2-axial image sets and resampled to 1 mm isotropic resolution. Normalization was performed independently for each modality and for each patient. Top 1% histogram was clipped for T2W to remove outlier pixels.

## 2.2 Data Augmentation

Classical data augmentation options include rotation, translational shifting, cropping, scaling, adding gaussian or salt-n-pepper noise, etc. We carefully selected rotation and scaling as our augmentation techniques while avoided adding irrelevant information that might bias the model's decision-making such as adding noise. A point indicating a lesion's location was provided in the original dataset for each lesion. With that point as the center, patches of 30x30, 32x32 and 34x34 pixels were cropped. The selection of patch size depended on lesion size for the whole data set and was verified as optimal by experiments.

The ratio of malignant versus benign lesions was around 1:3, and this was taken into consideration when performing image rotations. Two augmentation policies were designed to balance the training dataset between malignant and benign lesions. In Policy one only patches of malignant lesions were rotated and rotation angle was restricted to [2°, 4°, 6°]. While in Policy two benign lesion patches were randomly rotated for 5 times and malignant lesion patches for 19 times.

## 2.3 CNN Model and the End-to-End Training

Considering the scale and characteristics of our data and as a result of testing on various depths (2-5 convolutional layers), widths (16, 32, or 64 feature maps) and kernel sizes (3x3, 4x4 or 5x5), we finally utilized a vanilla CNN of VGG style[13,15] that consists of four convolutional layers, each followed by batch normalization and Relu nonlinear layers, and two max pooling layers after every two convolutional blocks.

The number of feature maps for each convolutional layer was 32, 32, 64, and 64, respectively. Fully connected layers consisted of two with 512 and 128 hidden nodes, respectively (Figure 2(a)).

The convolutional kernels underwent an orthogonalization process after being randomly initialized, which enforced diversity in learned features (Figure 3). Otherwise, there would be unlearned filters exhibiting blank feature maps. Adam solver was used for optimization. We used 10-fold cross validation in search for optimal meta parameters. The models with the best validation accuracy was then selected to make prediction on the testing cohort.

(a) Convolutional Neural Network Structure

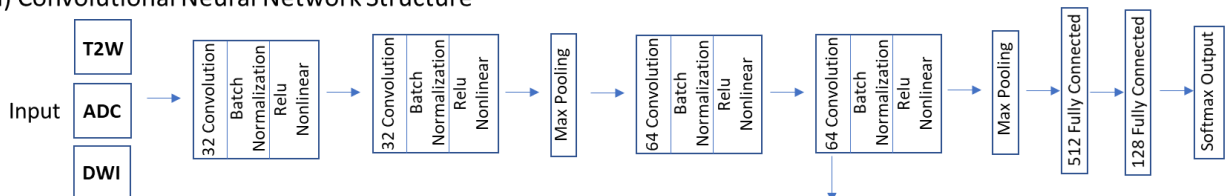

(b) Class Activation Map
- Network Training

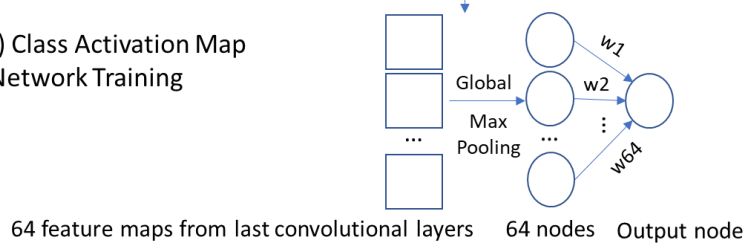

64 feature maps from last convolutional layers    64 nodes    Output node

(c) Class Activation Map
- Map Calculation

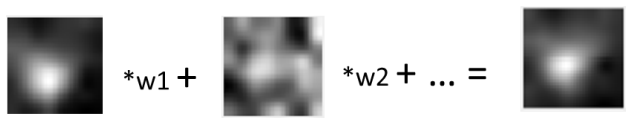

Figure 2. (a) Illustration of the Convolutional neural network (CNN) structure. (b) To calculate the class activation map (CAM), pre-trained CNN was truncated until the last convolutional layer and concatenated with a global pooling layer and output node. This new network was then fine-tuned based on parameters trained at stage (a). (c) The CAM was calculated as a linear combination of feature maps from the last convolutional layer. The weighting of each feature map was learned in stage (b).

## 2.4 Model Interpretation

To explore the impact of each individual modality, we trained different models using various combinations of imaging modalities and extracted the intermediate feature map for visualization at each combination.

We calculated the class activation map (CAM) [15] to find out the salient region on the input image patch. The CAM training process (Figure 2(b)) included down-sampling each feature map of the last convolutional layer into a node by global max pooling, removing the fully connected layers while keeping the output nodes and fine-tuning the new network until convergence. The ultimate feature map was computed as a linear combination of the feature maps from the last convolutional layer (Figure 2(c)). The weighting factor of each feature map was obtained as the connecting weights between the same feature map and the output node. Therefore, the CAM indicates class-related salient locations within the input patch when the model makes predictions.

## 2.5 Weighted ELM Classifier

ELM was proposed as a substitute for backpropagation (BP) to train a single hidden layer feedforward neural network. In contrast to other popular classification algorithms such as BP-based neural networks or support vector machines (SVMs), solution of ELM was analytically computed instead of iteratively tuned, which makes its implementation outstandingly efficient. A kernelized version of ELM was used in our work. Specifically, a radial basis function (RBF) kernel ($K(\mathbf{u}, \mathbf{v}) = \exp(-\gamma \|\mathbf{u} - \mathbf{v}\|^2)$ with kernel parameter $\gamma$ to be tuned) was used to transform the data and then parameters of the model were computed as a solution to an optimization problem (Equation 1) that aimed to minimize the training errors, as well as the L-2 norm of the parameters to prevent overfitting[16],

$$Minimize: L_{wELM} = \frac{1}{2}\|\beta\|^2 + C\frac{1}{2}\sum_{i=1}^{N} w_i \epsilon_i^2 \qquad (1)$$

$$Subject\ to: h(x_i)\beta = t_i - \epsilon_i, \quad i = 1, \dots, N,$$

where $N$ is the number of samples, $\beta$ is the weight connecting the hidden and output layers to be analytically computed, $h(x_i)$ is the hidden layer mapping of input $x_i$, $\epsilon_i$ is the error from predicted output $(h(x_i)\beta)$ and the targeted output $t_i$, $C$ is the tunable parameter as a trade-off between training error and L2-norm, and $w_i$ is the weighting assigned to input $x_i$ when computing the loss function. In our experiments, $w_i$ was set inversely proportional to the number of samples in the class $x_i$ belonged to. Here, kernel is defined as the multiplication of two mapping functions: $K(x_i, x_j) = h(x_i) \cdot h(x_j)$.

# 3. Results and Discussion

## 3.1 Multi-Modality Selection

Cross validation (CV) was stratified to evenly extract samples from both categories. Table 1 shows validation results on one of the ten folds with various combinations of modalities as input. When a single MR modality was used as the input, functional imaging such as ADC and $DWI_{b50}$ outperformed morphologic imaging (i.e., T2W). Note that by using ADC alone, the model performance was improved slightly compared to using both T2W and ADC. ADC was found to have better classification performance than $DWI_{b50}$ when either single or multiple modalities were used as inputs. A combination of T2W, ADC and $DWI_{b50}$ was shown to have the best performance, where the sensitivity, specificity and G-mean were (1.00, 0.83, 0.91) respectively. G-mean is the square root of product of sensitivity and specificity as a metric commonly used for imbalanced data. The performance was deteriorated when $k^{trans}$ was added to T2W, ADC and $DWI_{b50}$ to train the model.

Table 1. Validation results on one-fold of data with different combinations of modalities as input. The combination of T2W, ADC, $DWI_{b50}$ achieves the best performance.

| CV7 modality combinations | T2W | ADC | $DWI_{b50}$ | T2W +ADC | T2W +$DWI_{b50}$ | ADC + $DWI_{b50}$ | T2W +ADC + $DWI_{b50}$ | T2W +ADC + $DWI_{b50}$ + $k^{trans}$ |
|---|---|---|---|---|---|---|---|---|
| Sensitivity | 0.57 | 0.71 | 0.71 | 0.71 | 0.43 | 0.86 | 1.00 | 0.71 |
| Specificity | 0.63 | 0.88 | 0.83 | 0.79 | 0.88 | 0.88 | 0.83 | 0.88 |
| G-mean | 0.60 | 0.79 | 0.77 | 0.75 | 0.62 | 0.87 | 0.91 | 0.79 |

## 3.2 Lesion Classification

The results of the validation dataset were similar between two augmentation policies. As shown in Table 2, for the CNN end-to-end training, the average validation result was (0.53, 0.83, 0.65). Noticeable variations were observed among the ten folds, which were attributed to the heterogeneous distribution of lesion size, location, correlation among different MR modalities etc. However, more samples and wider rotation range proved with better generalization performance on the testing cohort 1 (Table 2). wELM was able to further improve the generalization performance as shown in Table 2.

ResNet50 was extensively validated with training data from fold CV7. The overloaded size of parameters caused overfitting with improved validation result but slight performance drop on testing set, compared with vanilla CNN.

Table 2. 10-fold cross validation and testing results. For each fold, there are about 7 malignant and 24 benign lesions. The first seven rows show the results of validation (top five rows) and test cohort 1 (row 6 & 7) using the end-to-end training. The last row is the results using CNN features with weighted ELM. AUC stands for arear under curve. Aug1(2) is an abbreviation for augmentation policy1(2).

| | | | | CV1 | CV2 | CV3 | CV4 | CV5 | CV6 | CV7 | CV7 ResNet 50 | CV8 | CV9 | CV10 |
|---|---|---|---|---|---|---|---|---|---|---|---|---|---|---|
| V A L I D A T I O N | End To End | Aug2 | Sensitivity | 0.67 | 0.43 | 0.71 | 0.29 | 0.43 | 0.57 | 1.00 | 1.00 | 0.43 | 0.29 | 0.46 |
| | | | Specificity | 0.88 | 0.92 | 0.88 | 0.79 | 0.83 | 0.83 | 0.83 | 0.96 | 0.79 | 0.75 | 0.82 |
| | | | G-mean | 0.77 | 0.63 | 0.79 | 0.48 | 0.60 | 0.69 | 0.91 | 0.98 | 0.58 | 0.47 | 0.61 |
| | | | AUC | 0.71 | 0.73 | 0.89 | 0.54 | 0.64 | 0.76 | 0.92 | 0.93 | 0.67 | 0.49 | 0.69 |
| | | | Accuracy | 0.81 | 0.80 | 0.85 | 0.67 | 0.68 | 0.78 | 0.85 | 0.87 | 0.71 | 0.64 | 0.72 |
| T E S T | End To End | Aug1 | AUC | 0.75 | 0.78 | 0.78 | 0.72 | 0.80 | 0.71 | 0.77 | - | 0.78 | 0.81 | 0.79 |
| | | Aug2 | AUC | 0.80 | 0.77 | 0.79 | 0.81 | 0.82 | 0.81 | 0.81 | 0.79 | 0.79 | 0.82 | 0.80 |
| | wELM | | AUC | 0.82 | 0.78 | 0.81 | 0.81 | 0.83 | 0.82 | 0.84 | - | 0.80 | 0.82 | 0.80 |

We further extracted features from each convolutional layer, fully connected layers, and the combination of the first and last convolutional layers and compared their discriminant powers using wELM (Table 3). The model trained in CV7 fold was investigated since it had the best performance in the previous test. The features extracted from the third convolutional layer (C3) or the combined C1+C4 had the best performance highlighted in Table 3. The (Sensitivity, specificity, G-mean) were increased from (1.00, 0.83, 0.91) to (1.00, 0.92, 0.96). Then we retrained the wELM model using 10-fold cross validation by feeding wELM the features extracted from C1+C4. Table 4 shows the model performance of validation set at each individual CV fold. The average performance using C1+C4 and wELM was (0.76, 0.82, 0.79) for (sensitivity, specific, G-mean), increasing from (0.53, 0.83, 0.65) achieved by the CNN end-to-end training.

Table 3. Validation results using the combination of features from various layers and weighted extreme learning machine (wELM) classifier based on the best performing cross validation fold CV7. By using features from the third convolutional layer (C3) or the combination of the first and the last convolutional layers (C1+C4), performance increased from (1, 0.83, 0.91) to (1, 0.92, 0.96), the same as two more benign lesions being correctly classified. C is an abbreviation for convolutional layer and FC for fully-connected layer.

| CV7 CNN features+ wELM | C1 | C2 | C3 | C4 | FC1 | FC2 | C1 +C4 |
|---|---|---|---|---|---|---|---|
| Sensitivity | 1.00 | 1.00 | 1.00 | 1.00 | 0.86 | 1.00 | 1.00 |
| Specificity | 0.88 | 0.79 | 0.92 | 0.83 | 0.88 | 0.83 | 0.92 |
| G-mean | 0.94 | 0.89 | 0.96 | 0.91 | 0.87 | 0.91 | 0.96 |

Table 4. 10-fold cross validation (CV) results using features from the first and the last convolutional layer (C1+C4) and weighted extreme learning machine (wELM) as the classifier. Compared to the CNN end-to-end training, the average performance increased from (0.53, 0.83, 0.65) to (0.76, 0.82, 0.79) for (sensitivity, specific, G-mean).

| CNN features+ wELM | CV1 | CV2 | CV3 | CV4 | CV5 | CV6 | CV7 | CV8 | CV9 | CV10 |
|---|---|---|---|---|---|---|---|---|---|---|
| Sensitivity | 1.00 | 0.71 | 1.00 | 0.57 | 0.71 | 0.57 | 1.00 | 0.71 | 0.71 | 0.62 |
| Specificity | 0.84 | 0.83 | 0.83 | 0.67 | 0.92 | 0.92 | 0.92 | 0.75 | 0.79 | 0.75 |
| G-mean | 0.92 | 0.77 | 0.91 | 0.62 | 0.81 | 0.72 | 0.96 | 0.73 | 0.75 | 0.68 |

The dataset we used was from the ProstateX Grand Challenge with the aim to differentiate between clinically significant and not significant lesions from mpMRI. Over the 71 computerized methods, the AUC of the testing cohort with 208 lesions ranged from 0.45 to 0.87[12]. Our model achieved an AUC of 0.84 which was in line with other top performing models.

We also applied the model to the testing cohort 2 collected at our institution with completely different scanner and scanning sequences. The CNN end-to-end model was found to generalize well with (0.87, 0.94, 0.90) for sensitivity, specific and G-mean, respectively. In cohort 2, patients were scanned with a scanner from a different vendor compared to cohort 1. The scanning parameters were different. Moreover, $DWIb_{1000}$ rather than $DWI_{b50}$ was acquired in cohort 2, which potentially had the largest impact on the classification results. Though our testing cohort size was small, the results implied that our model had better capacity to address the domain shifts when using cross-institutional data sources.

### 3.3 Feature Mapping Visualization

We showed that the imaging sequences had strong impact on the classification accuracy. The feature maps learned from each convolutional layer using different combinations of MR sequences as input were discussed and illustrated using the case shown in Figure 3. The best results were achieved using a combination of T2W, $DWI_{b50}$, and ADC. As shown in Figure 3(a), through visualization of deep feature maps after each convolutional layer, the lower layer was able to capture the edges of each structure including the prostate, rectum and obturator internus muscles, and the sub-regions inside the prostate including the CG and PZ. The feature map also covered a spectrum of intensity changes contributed from each image modality. As it moved to the deep layers, the spatial resolution of the feature maps was reduced and focused on the abnormal regions where the lesion resided, and the learned features tended to be more abstract and task related. As shown in the feature maps of the last layer, the lesion (blue) was visible in a majority of the feature maps and a hypointense area (red) in the lesion's close proximity was also included in a subset of feature maps. These feature maps were passed to the fully connected layer and used for classification. This observation was consistent with that on natural color images[17].

(a)

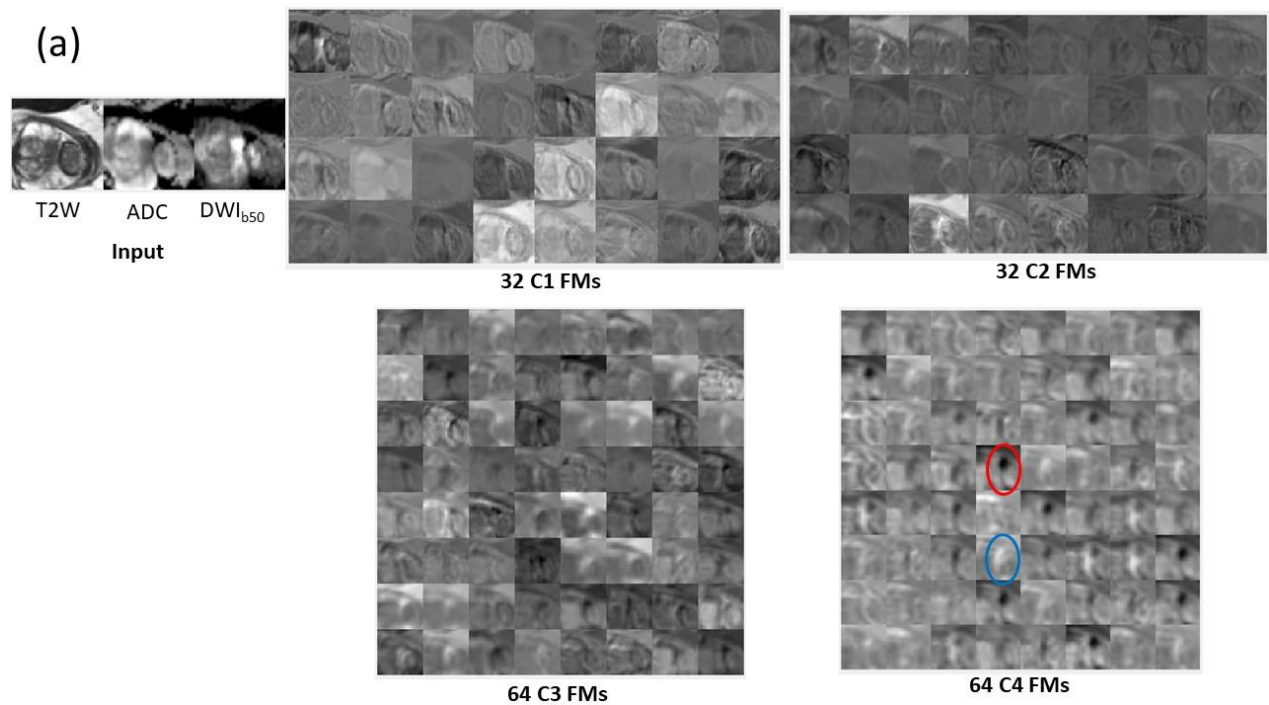

(b)

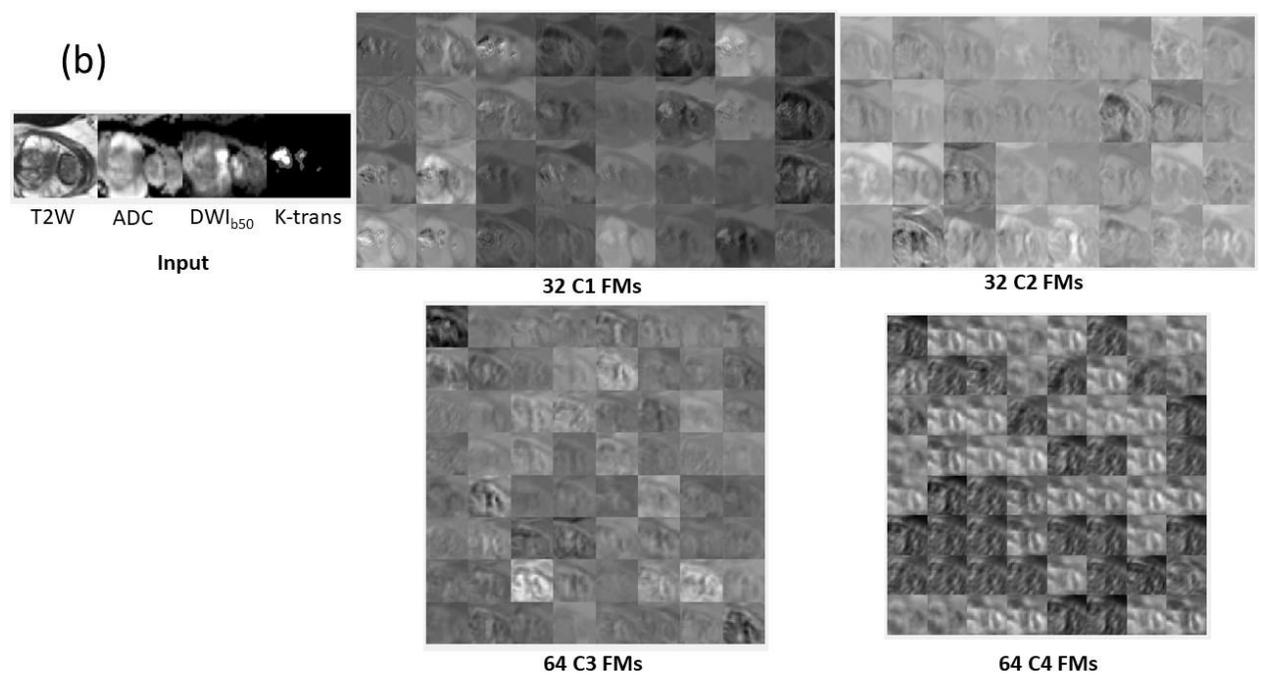

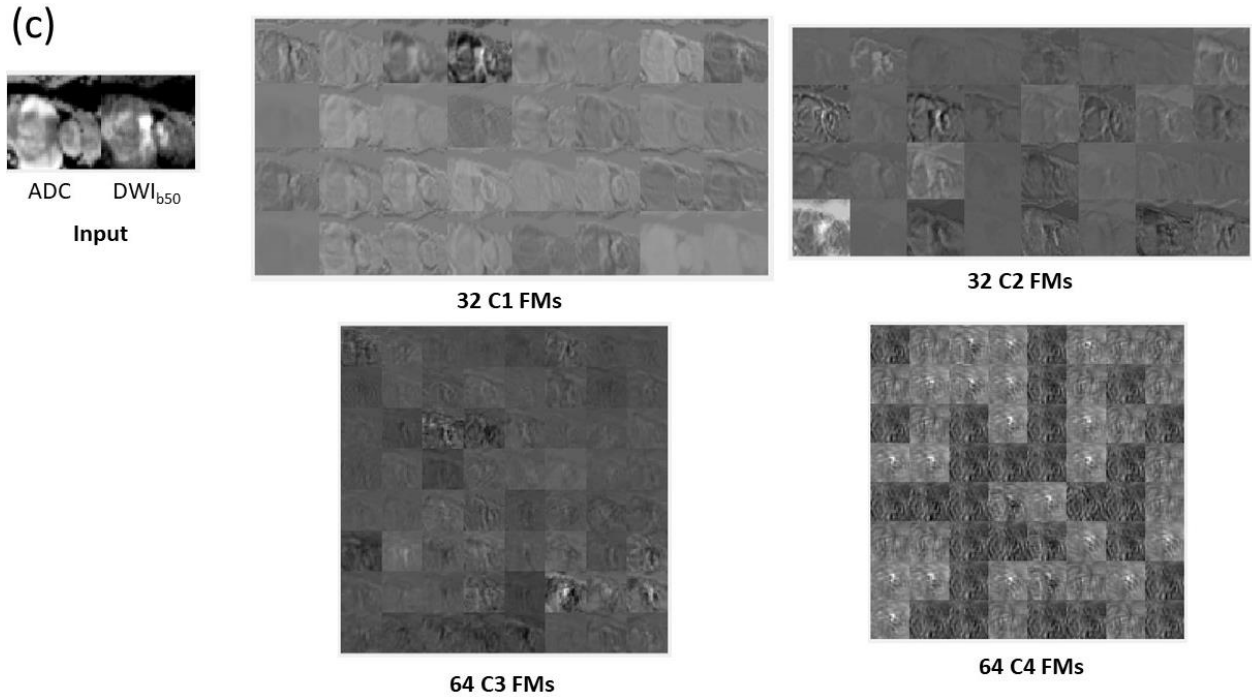

Figure 3: Visualization of intermediate feature mappings for every convolutional layer with respect to a malignant lesion in the peripheral zone (PZ). (a) Model trained with T2, ADC and $DWI_{b50}$ as multi-channel inputs successfully detected the lesion as malignant; (b) model with T2, ADC, $DWI_{b50}$ and $k^{trans}$ as input; (c) and model with ADC and $DWI_{b50}$ as input both failed. Moreover, (a) exhibits more rich and hierarchical features than (b) and (c). C1 stands for convolution layer 1, similarly for C2, C3, C4 as well. FMs is abbreviation for feature mappings.

The combination of T2W, DWI, and DCE has been shown to provide high sensitivity and specificity in intraprostatic lesion identification and has been recommended for lesion detection in current consensus guidelines[18]. We also investigated why our model performance was degraded by adding $k^{trans}$. As shown in Figure4(b), the $k^{trans}$ map was added as an additional input. There were two regions elevated in the $k^{trans}$ map, a brighter area in the CG and a relatively less intense area in the lesion. $K^{trans}$ does not differentiate PCa from benign prostatic hyperplasia reliably within the CG due to similar microvascular density exhibited in both conditions. By incorporating this information in the learning, the model was able to capture additional information from the $k^{trans}$ of the first layer. However, we observed that the feature maps shifted toward the CG which was shown in the last layer. The lesion was observable in most of the feature maps. It clearly induced more uncertainty in the classification task and impacted model performance.

DWI has been shown to have good sensitivity and specificity for identification of PCa. We repeated a similar procedure to learn how the features were learned using ADC and $DWI_{b50}$ alone (Figure 3(c)). Considering that DWI measures water molecules' diffusion behavior, it lacks anatomical and structural information important for identifying the edges and location of each structure. Hence, in the first layer, a portion of feature maps missed detailed information and were hardly visible. These feature maps were propagated to deeper layers and created very noisy maps. The lesion (hypointense on ADC and hyperintense on $DWI_{b50}$) was captured in a majority of feature maps in the last layer. It was difficult to make classifications based on these feature maps considering reduced features of the shape, location etc.

We investigated the model performance on lesion classification based on different combinations of MR sequences as input. By investigating the feature maps at each layer, it gained better understanding of how the CNN model was trained, and this information can significantly contribute developing a better strategy for lesion classification using MR images. Clearly, T2W provides morphological features important in identifying lesion shapes and locations. Considering the high sensitivity and specificity values, DWI and ADC are critical imaging modalities in PCa detection. DWI plus T2W was significantly more sensitive and accurate than T2W alone in tumor volume measurement accuracy[19]. Kim et al.[20] showed that the combination of T2W and DWI improved the area under receiver operating characteristics curve (AUC) from 0.61 to 0.88 under clinical interpretation. The addition of DWI to T2W improved PCa localization performance, with AUC increased from 0.66 to 0.79 in the prospective studies[21]. Here, we achieved an excellent classification accuracy using the combination of T2W and DWI. Though $k^{trans}$ has been shown to improve the lesion detection in the clinical setting, caution need to be taken when integrating $k^{trans}$ into training. As discussed earlier, the elevated $k^{trans}$ values were present in both lesions and other areas with similar microvascular conditions such as benign prostatic hyperplasia, or vascular enlargement in the gland. It induced higher false positive rates in the model. A potential solution is to split the data into CG and PZ and to train the model separately based on the anatomical structures. This strategy needs further investigation.

Figure 4 shows the T-sne plots using T2W, ADC and DWI$_{b50}$ as inputs. The features extracted from the entire patient population from C1, C4 and FC1 were projected into 2-dimensional space to reveal the trend of distinguishability between malignant and benign lesions as the CNN network layer propagated.

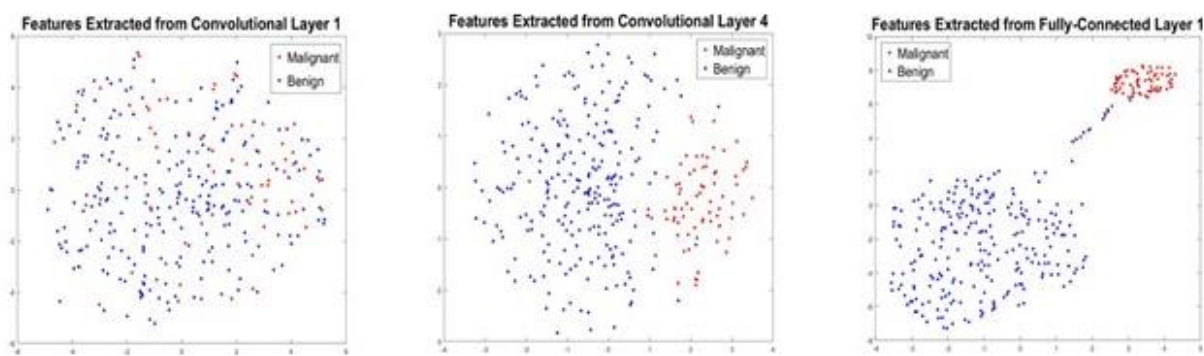

Figure 4: With T2, ADC and DWI$_{b50}$ as inputs, T-sne plots of features for the entire population extracted from different layers (C1, C4 and FC1). As the layer propagated, the features became more distinguishable between the malignant and benign lesions. C1 stands for convolution layer 1, similarly for C2, C3, C4 as well. FMs is abbreviation for feature mappings.

In Figure 5, we presented the global average feature maps at each convolutional layer for 12 patients including the one presented in Figure 3. It showed that C1 was able to detect all the edges between pelvic and background, bladder, and prostate, prostate and rectum, central gland and peripheral zone, lesion and surrounding normal tissues. C2 showed hyperintense signals for each organ including prostate, bladder, and rectum. C3 narrowed down further by looking into the heterogeneity of prostate gland and detecting the abnormality. Finally, C4 removed irrelevant information and focused only on the suspicious area with hypointense signals. Global feature maps calculated from C4 (green) were fused with T2W (pink) to show that suspicious regions derived from the global feature map aligned well (brown) with the abnormal regions visualized in T2W.

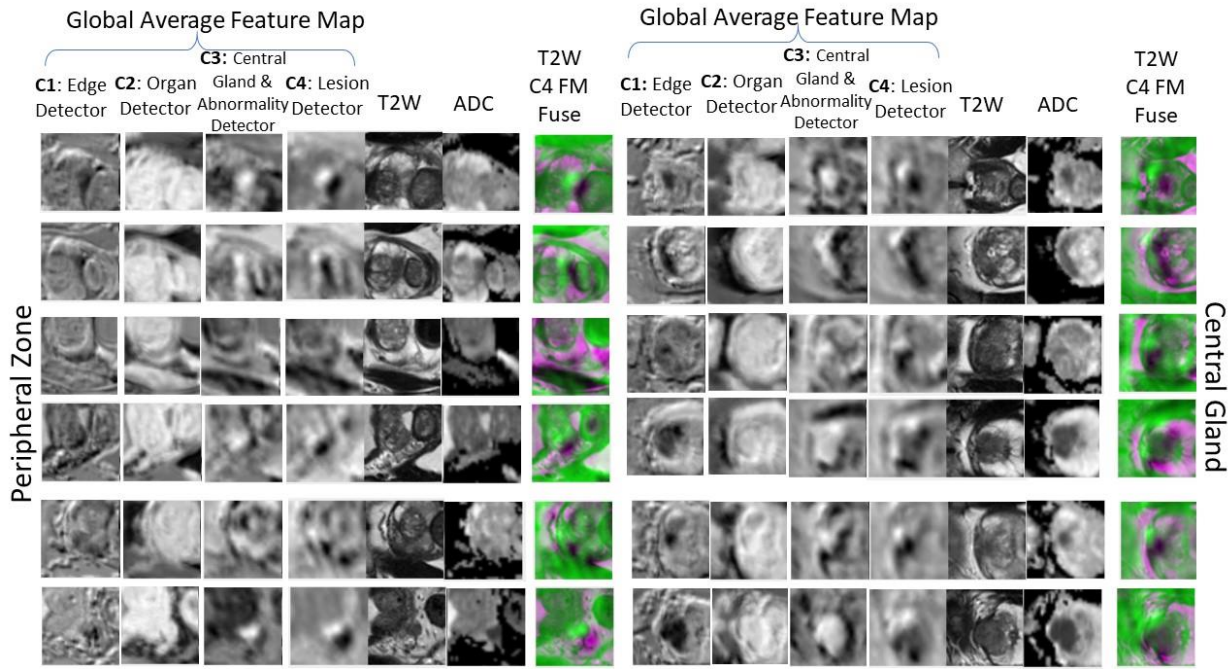

Figure 5: Average of all feature maps extracted from each convolutional layer for 12 randomly selected lesions.

## 3.4 Saliency Map

The CAM provides important clues on how the deep learning models make decisions. Figure 6 (a) displayed the CAM calculated based on the 34x34 pixel patches from Figure 1 with two malignant lesions in PZ, and two benign cases in CG. When the data was augmented with limited rotation angles, the model was found with a tendency to focus on a narrower region-of-interest in the suspicious area (CAM1 in Figure 6), which could lead to false predictions. On the other hand, by rotating images with a wider range, the variability of the background including the normal prostatic gland, was strengthened. The resultant model was found with broader attention window (hyperintense signals) around the suspicious area (Figure 6 CAM2), which mimic clinicians' practices on MRI reading by comparing the contrast of suspicious lesions to the surrounding tissues. This may explain its superior performance on the testing cohort.

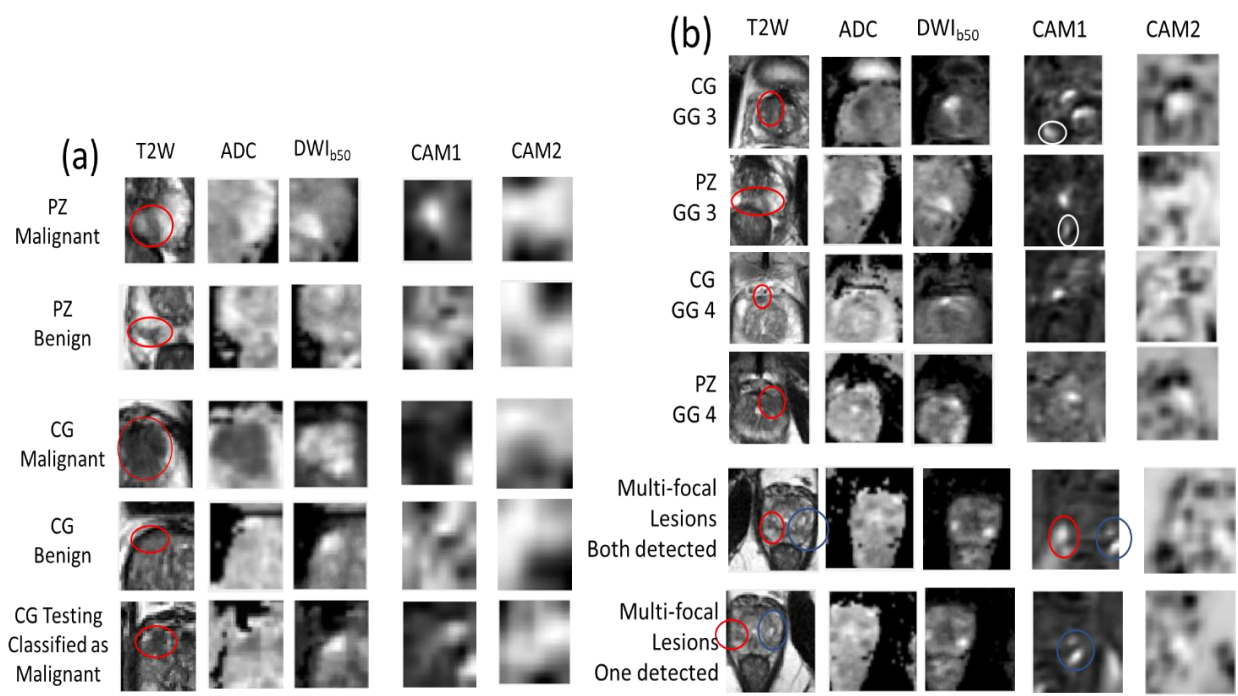

Figure 6. (a) The CAM of cases presented in Figure 1 with 34x34 pixel patch input. The CAM1 and CAM2 were generated by models trained with data augmentation policy 1 and 2, respectively. (b) Filters learned from 34x34 pixel patch augmented under Policy 1 can generalize well to 64x64 pixel patch, where all lesions were exclusively detected in most cases. The computed saliency maps were found overlay well with lesions, thus having potential for real clinical applications. Abbreviations: GG – Gleason grade Group; PZ – peripheral zone; CG – central gland.

We then tested on 64x64 pixel patches to see if filters learned from small patch size can generalize to cases with a bigger patch size. As can be seen from Figure 6 (b), the CNN model with augmentation policy one was able to detect lesions exclusively within the prostate and exclude adjacent organs such as the bladder and rectum. For the first two cases, the prostate boundary (white circled) was also highlighted. The fifth and sixth cases showed patients with two lesions in the PZ, circled in blue and red, respectively. In the fifth case, both lesions were successfully detected and located; while in the sixth case, the lesion circled in red failed to be detected, mainly because it was cut off in the input patch.

This indicated potential application of using CAM for intraprostatic lesion delineation for diagnostic and therapeutic purposes.

# 4. Conclusions

This work developed a framework to accurately classify prostate lesion malignancy from mpMRI. We demonstrated that the combination of feature extraction using the deep learning model and the weighted extreme machine classification algorithm was an effective solution to address the small and imbalanced sample size challenges. Based on both qualitative and quantitative evaluation, the combination of T2W, ADC and DWI were considered as the optimal multi-channel inputs. The feature map visualization and saliency map can provide additional insights to augment clinicians in localization of suspicious regions-of-interest. Another important finding was the generalization capability of a CNN model with appropriate data augmentation when trained and tested on mpMRI from different scanners or acquired with different protocols, which will be further investigated in the future work.

**Acknowledgement**

This work was supported by a Research Scholar Grant, RSG-15-137-01-CCE from the American Cancer Society.